\PassOptionsToPackage{unicode}{hyperref}
\PassOptionsToPackage{hyphens}{url}
\PassOptionsToPackage{dvipsnames,svgnames,x11names}{xcolor}
\documentclass[
]{article}

\usepackage{xcolor}
\usepackage{amsmath,amssymb}

\usepackage{fontspec}
\usepackage[numbers,sort&compress]{natbib}
\IfFileExists{upquote.sty}{\usepackage{upquote}}{}
\IfFileExists{microtype.sty}{%
  \usepackage[]{microtype}
  \UseMicrotypeSet[protrusion]{basicmath}
}{}

\makeatletter
\@ifundefined{KOMAClassName}{%
  \IfFileExists{parskip.sty}{%
    \usepackage{parskip}
  }{%
    \setlength{\parindent}{0pt}
    \setlength{\parskip}{6pt plus 2pt minus 1pt}}
}{%
  \KOMAoptions{parskip = half}}
\makeatother

\makeatletter
\ifx\paragraph\undefined\else
  \let\oldparagraph\paragraph
  \renewcommand{\paragraph}{
    \@ifstar
      \xxxParagraphStar
      \xxxParagraphNoStar
  }
  \newcommand{\xxxParagraphStar}[1]{\oldparagraph*{#1}\mbox{}}
  \newcommand{\xxxParagraphNoStar}[1]{\oldparagraph{#1}\mbox{}}
\fi
\ifx\subparagraph\undefined\else
  \let\oldsubparagraph\subparagraph
  \renewcommand{\subparagraph}{
    \@ifstar
      \xxxSubParagraphStar
      \xxxSubParagraphNoStar
  }
  \newcommand{\xxxSubParagraphStar}[1]{\oldsubparagraph*{#1}\mbox{}}
  \newcommand{\xxxSubParagraphNoStar}[1]{\oldsubparagraph{#1}\mbox{}}
\fi
\makeatother

\usepackage{color}
\usepackage{fancyvrb}

\DefineVerbatimEnvironment{Highlighting}{Verbatim}{commandchars = \\\{\}}
\usepackage{framed}
\definecolor{shadecolor}{RGB}{241,243,245}
\newenvironment{Shaded}{\begin{snugshade}}{\end{snugshade}}

\newcommand{\CommentTok}[1]{\textcolor[rgb]{0.37,0.37,0.37}{#1}}

\newcommand{\ControlFlowTok}[1]{\textcolor[rgb]{0.00,0.23,0.31}{\textbf{#1}}}

\newcommand{\DecValTok}[1]{\textcolor[rgb]{0.68,0.00,0.00}{#1}}

\newcommand{\FloatTok}[1]{\textcolor[rgb]{0.68,0.00,0.00}{#1}}

\newcommand{\KeywordTok}[1]{\textcolor[rgb]{0.00,0.23,0.31}{\textbf{#1}}}
\newcommand{\NormalTok}[1]{\textcolor[rgb]{0.00,0.23,0.31}{#1}}
\newcommand{\OperatorTok}[1]{\textcolor[rgb]{0.37,0.37,0.37}{#1}}

\newcommand{\StringTok}[1]{\textcolor[rgb]{0.13,0.47,0.30}{#1}}
\newcommand{\VariableTok}[1]{\textcolor[rgb]{0.07,0.07,0.07}{#1}}

\usepackage{longtable,booktabs,array}
\usepackage{calc}
\usepackage{etoolbox}
\makeatletter
\patchcmd\longtable{\par}{\if@noskipsec\mbox{}\fi\par}{}{}
\makeatother
\IfFileExists{footnotehyper.sty}{\usepackage{footnotehyper}}{\usepackage{footnote}}
\makesavenoteenv{longtable}

\usepackage{graphicx}
\makeatletter
\newsavebox\pandoc@box
\newcommand*\pandocbounded[1]{%
  \sbox\pandoc@box{#1}%
  \Gscale@div\@tempa{\textheight}{\dimexpr\ht\pandoc@box+\dp\pandoc@box\relax}%
  \Gscale@div\@tempb{\linewidth}{\wd\pandoc@box}%
  \ifdim\@tempb\p@<\@tempa\p@\let\@tempa\@tempb\fi%
  \ifdim\@tempa\p@<\p@\scalebox{\@tempa}{\usebox\pandoc@box}%
  \else\usebox{\pandoc@box}%
  \fi%
}
\def\fps@figure{htbp}
\makeatother

\providecommand{\tightlist}{%
  \setlength{\itemsep}{0pt}\setlength{\parskip}{0pt}}

\makeatletter
\@ifpackageloaded{caption}{}{\usepackage{caption}}
\AtBeginDocument{%
\ifdefined\contentsname
  \renewcommand*\contentsname{Table of contents}
\else
  \newcommand\contentsname{Table of contents}
\fi
\ifdefined\listfigurename
  \renewcommand*\listfigurename{List of Figures}
\else
  \newcommand\listfigurename{List of Figures}
\fi
\ifdefined\listtablename
  \renewcommand*\listtablename{List of Tables}
\else
  \newcommand\listtablename{List of Tables}
\fi
\ifdefined\figurename
  \renewcommand*\figurename{Figure}
\else
  \newcommand\figurename{Figure}
\fi
\ifdefined\tablename
  \renewcommand*\tablename{Table}
\else
  \newcommand\tablename{Table}
\fi
}
\@ifpackageloaded{float}{}{\usepackage{float}}
\floatstyle{ruled}
\@ifundefined{c@chapter}{\newfloat{codelisting}{h}{lop}}{\newfloat{codelisting}{h}{lop}[chapter]}
\floatname{codelisting}{Listing}

\makeatother
\makeatletter
\@ifpackageloaded{caption}{}{\usepackage{caption}}
\@ifpackageloaded{subcaption}{}{\usepackage{subcaption}}
\makeatother

\usepackage{bookmark}
\IfFileExists{xurl.sty}{\usepackage{xurl}}{}
\hypersetup{
  pdftitle = {Hot Hẻm: Sài Gòn Giữa Cái Nóng Hổng Công Bằng---Saigon in Unequal Heat},
  pdfauthor = {Tess Vu},
  pdfsubject = {cs.CV, cs.LG, cs.CY, physics.soc-ph},
  pdfkeywords = {GeoAI, Ho Chi Minh City, Saigon, Vietnam, XGBoost, data science, environmental justice, geospatial, machine learning, pedestrian routing, spatial, street view imagery, thermal mapping, urban analytics, urban heat island},
  colorlinks = true,
  linkcolor = {blue},
  filecolor = {Maroon},
  citecolor = {Blue},
  urlcolor = {Blue},
  pdfcreator = {LaTeX via pandoc}}

\title{Hot Hẻm: Sài Gòn Giữa Cái Nóng Hổng Công Bằng---Saigon in Unequal Heat\\
\large Optimization for Suffering: The Hottest Route Available}

\author{Tess Vu\\
\textit{Master of Urban Spatial Analytics}\\
Stuart Weitzman School of Design\\
University of Pennsylvania\\
Philadelphia, PA 19104, USA\\
\texttt{tessavu@upenn.edu}}

\date{December 9, 2025}

\begin{document}
\maketitle

\begin{abstract}
Pedestrian heat exposure is a critical health risk in dense tropical cities, yet standard routing algorithms often ignore micro-scale thermal variation. Hot Hẻm is a GeoAI workflow that estimates and operationalizes pedestrian heat exposure in Hồ Chí Minh City (HCMC), Việt Nam, colloquially known as Sài Gòn. This spatial data science pipeline combines Google Street View (GSV) imagery, semantic image segmentation, and remote sensing. Two XGBoost models are trained to predict land surface temperature (LST) using a GSV training dataset in selected administrative wards, known as phường, and are deployed in a patchwork manner across all OSMnx-derived pedestrian network nodes to enable heat-aware routing. This is a model that, when deployed, can provide a foundation for \emph{pinpointing where} and further \emph{understanding why} certain city corridors may experience disproportionately higher temperatures at an infrastructural scale.
\end{abstract}


\section{Introduction}
\label{sec:introduction}

Given that urban heat and other environmental injustices are widely recognized as being disproportionately felt \citep{chakraborty2019}, dangerous heat exposure will only continue to exacerbate with growing populations and current pollution trajectories. Extreme heat is heterogeneous and driven by both macro-scale morphologies (e.g., elevation, land cover, surface emissivity) and micro-scale streetscapes (e.g., building canyon effects, tree canopy, visible sky) \citep{oke1982}, many of which are influenced by local municipalities' regulations on the built environment and social structures.

Conventional thermal mapping often emphasizes satellite-derived patterns that could underrepresent pedestrian-scale experiences \citep{middel2019}, and some existing literature notes shaded routes can significantly improve pedestrian comfort. However, there is lacking emphasis that the onus falls on local municipalities to provide resilient, cool, and green infrastructure---this is a byproduct communicated by shade-finding algorithms that present coolest routes. Regardless of intention, they present as alternatives rather than tools to assist with building solutions, implying health, wellbeing, and heat-stress mitigation is a choice among locals, and not a prevailing systemic and infrastructural issue that will worsen with global warming.

This project aims to fill these gaps by firstly, fusing street-level visual morphology with thermal and structural remote-sensing predictors, and secondly, by seeking the hottest routes as a government tool. This is where machine learning (ML) optimization can recommend routes \emph{minimizing shade} and \emph{maximizing sun exposure}, revealing the hottest paths as potential candidates for shaded infrastructure, future tree canopies, or further investigation, demonstrating how ML can help enhance urban resilience to extreme heat.

\subsection{Related Work}
\label{subsec:related-work}

This work builds on three converging research streams: urban thermal remote sensing, street-level imagery analytics, and heat-aware pedestrian routing.

\subsubsection{Urban Heat Island and Thermal Remote Sensing}
\label{subsubsec:uhi}

The urban heat island (UHI) effect---where cities experience elevated temperatures relative to surrounding rural areas---has been extensively documented since Oke's work on urban energy balance \citep{oke1982}. Satellite-based thermal remote sensing enables city-scale LST mapping \citep{voogt2003}, though the coarse spatial resolution (30m for Landsat) limits representation of micro-scale thermal variation experienced by pedestrians \citep{ho2014}. Recent work has shown that lower-income neighborhoods experience disproportionately higher heat exposure \citep{chakraborty2019}, emphasizing the environmental justice dimensions of urban heat.

\subsubsection{Street View Imagery for Urban Analytics}
\label{subsubsec:street-view}

GSV and similar platforms have provided unprecedented human-scale urban measurement. \citet{li2015} pioneered the Green View Index (GVI) to quantify street-level vegetation from GSV imagery. Subsequent work applied deep learning to extract urban morphology features including sky view, building density, and streetscape perception \citep{middel2019, zhang2019}. Comprehensive reviews by \citet{kang2020} and \citet{biljecki2021} document the expanding role of street view imagery in public health and urban analytics, although applications to thermal comfort prediction remain limited.

\subsubsection{Heat-Aware Routing}
\label{subsubsec:routing}

While shortest-path algorithms like Dijkstra's \citep{dijkstra1959} are well-established, incorporating thermal comfort into routing optimization is relatively recent. Existing approaches typically seek coolest routes to minimize pedestrian heat exposure. This work inverts that framing: by identifying the \emph{hottest} routes, municipalities can be provided with actionable infrastructure priorities rather than placing the burden of heat avoidance on individuals.

\section{Data and Study Area}
\label{sec:data}

\subsection{Pedestrian Network and Wards}
\label{subsec:network}

A pedestrian network graph with a 500-meter buffer was extracted from Python's \texttt{osmnx} \citep{boeing2017}, yielding 28,445 nodes and 74,710 edges for three administrative districts: District 1, District 2, and District 8. Due to API costs and keeping in mind computational efficiency, only six wards were selected as disparate GSV candidates, two from each district of interest: Bến Thành (2,424 nodes) and Cô Giang (1,840 nodes) in District 1, An Khánh (1,119 nodes) and Thảo Điền (1,547) in District 2, and Phường 5 (2,346 nodes) and Phường 6 (2,013 nodes) from District 8.

\begin{figure}[H]
{\centering \pandocbounded{\includegraphics[keepaspectratio]{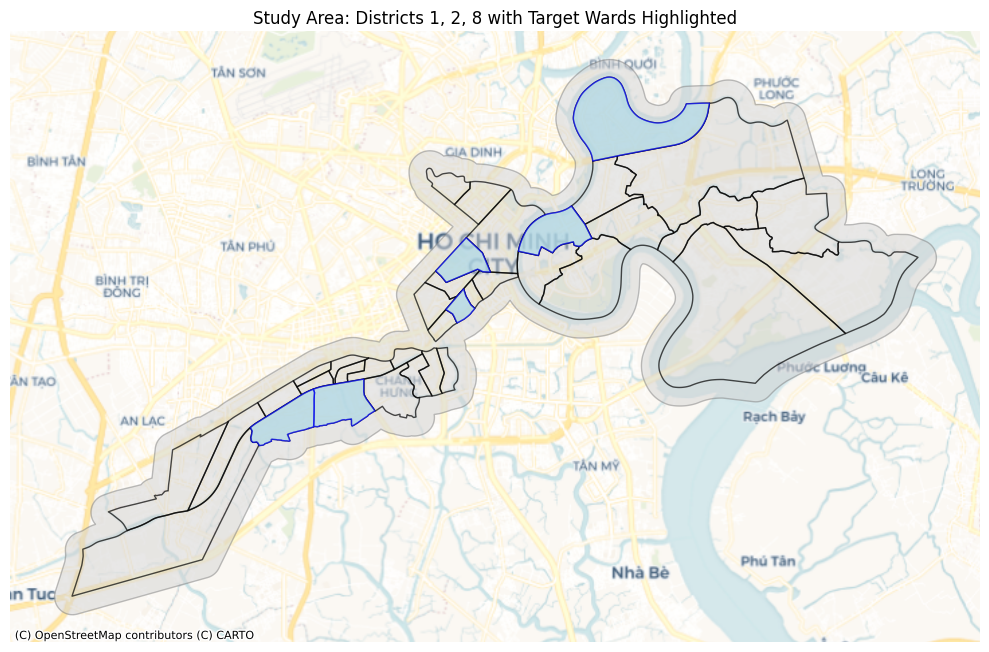}}
}
\caption{Study area showing the three administrative districts in Ho Chi Minh City.}
\label{fig:study-area}
\end{figure}

\begin{figure}[H]
{\centering \pandocbounded{\includegraphics[keepaspectratio]{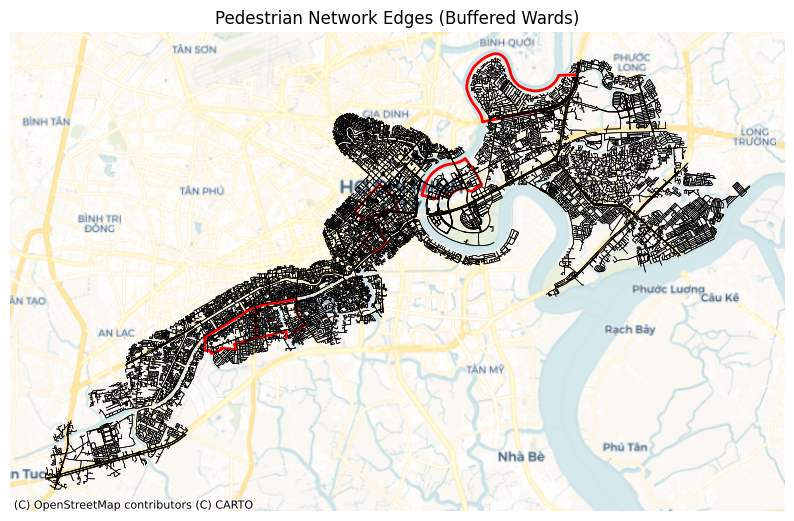}}
}
\caption{Pedestrian network extracted from OpenStreetMap covering the study area.}
\label{fig:network-area}
\end{figure}

\subsection{Street-Level Imagery}
\label{subsec:gsv}

GSV samples generated 23,806 points from 500-meter buffered wards of interest at 50-meter intervals. Metadata contained 20,457 images, an unfortunate 14.07\% decrease due to interruptions, but nonetheless providing sufficient density for training streetscape indices and validating segmentation outputs within wards of interest.

\subsection{Remote Sensing Rasters}
\label{subsec:rasters}

The satellite data was extracted from several different sources \citep{usgs2024, shimada2014}:

\textbf{Landsat 8 / 9 (30m resolution, dry months December thru April, 2023--2025)}

\begin{itemize}
\item LST ST\_B10 Band
\item Emissivity ST\_EMIS Band
\item Red SR\_B4 Band
\item Near Infrared (NIR) SR\_B5 Band
\item QA\_PIXEL Band
\end{itemize}

\textbf{JAXA LULC (10m resolution, 2025)}

\begin{itemize}
\tightlist
\item Land Cover
\end{itemize}

\textbf{JAXA PALSAR-2 (ScanSAR, 50m resolution, 2025)}

\begin{itemize}
\item HH Polarization
\item HV Polarization
\item Observation Date / Time
\item Local Incidence Angle
\item Mask / Flag
\end{itemize}

\textbf{ALOS World 3D DSM (30m resolution, 2025)}

\begin{itemize}
\item Elevation
\item Mask
\item Stacking Number
\end{itemize}

\section{Methodologies}
\label{sec:methods}

In the interest of computational efficiency, all notebooks and scripting were offloaded from local into cloud computing using Google's Colab Pro with A100 GPU acceleration. Combining Landsat rasters into a composite was completed in ArcGIS v3.4.

\subsection{Image Segmentation and Superclass Mapping}
\label{subsec:segmentation}

GSV images were processed using Mask2Former Swin-Large \citep{cheng2022} trained on Mapillary Vistas (\texttt{facebook/mask2former-swin-large-mapillary-vistas-semantic}) \citep{neuhold2017}, accessed via Hugging Face. The model contains 65 object categories optimized for street-level scene analysis in complex urban environments like Sài Gòn.

GSV images were downloaded at 640 × 640 pixel resolution. Ideally, panoramic images would have been preferable, but due to cost restraints, the static imagery was made to dynamically alter the header to be front-facing.

To improve interpretability and stability for index construction, raw segmentation classes were remapped into seven superclasses:

\begin{longtable}[]{@{}ll@{}}
\caption{Superclass mappings for semantic segmentation output.}\label{tab:superclass}\\
\toprule\noalign{}
Mapping & Superclass \\
\midrule\noalign{}
\endfirsthead
\toprule\noalign{}
Mapping & Superclass \\
\midrule\noalign{}
\endhead
\bottomrule\noalign{}
\endlastfoot
0 & Other \\
1 & Vegetation \\
2 & Sky \\
3 & Building \\
4 & Pavement / Road \\
5 & Water \\
6 & Vehicle / Street Clutter \\
\end{longtable}

The mapping aggregates the 65 Mapillary Vistas classes as follows: Other (23 classes including persons, animals, terrain, street furniture), Vegetation (1 class), Sky (1 class), Building (7 classes including walls, fences, bridges, tunnels), Pavement/Road (12 classes including sidewalks, bike lanes, parking), Water (2 classes including boats), and Vehicle/Clutter (16 classes including poles, signs, vehicles).

\begin{figure}[H]
{\centering \pandocbounded{\includegraphics[keepaspectratio]{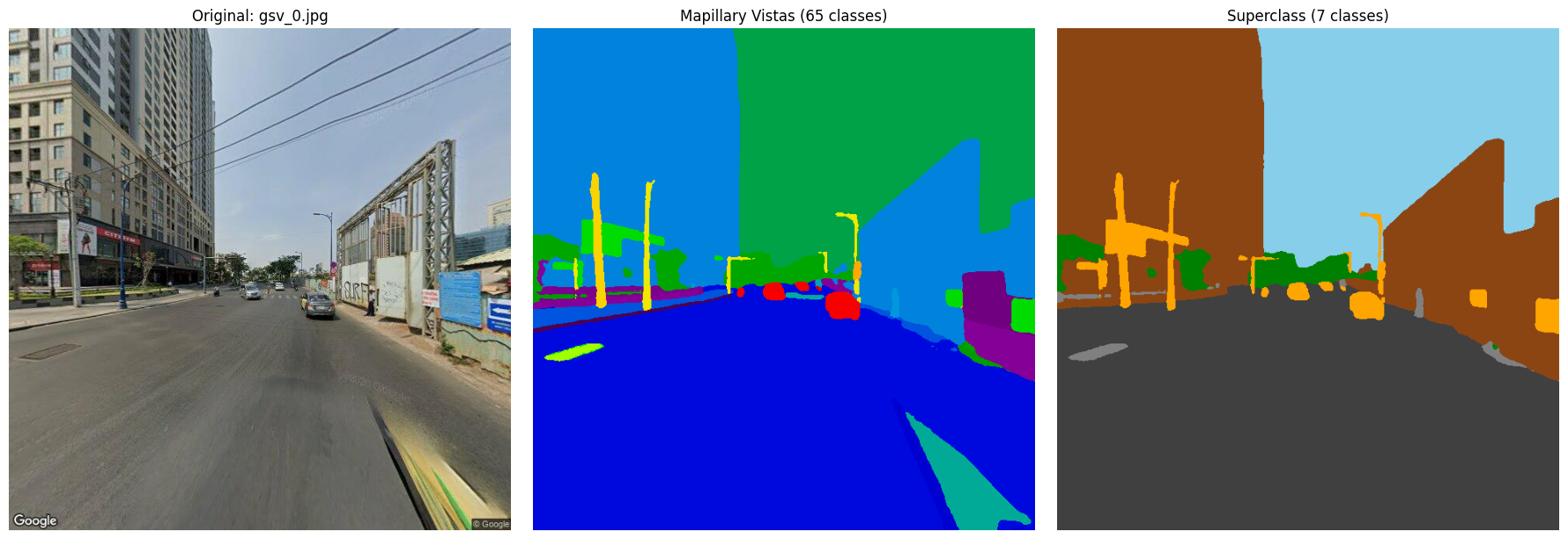}}
}
\caption{Example of semantic segmentation process applied to Google Street View imagery.}
\label{fig:segmentation}
\end{figure}

\subsection{Engineering Indices}
\label{subsec:indices}

The final merged superclass imagery was used to create predictive features \citep{li2015}. Seven superclass percentages were computed directly from the segmentation masks:

\begin{itemize}
\item \texttt{pct\_vegetation}: Proportion of vegetation pixels
\item \texttt{pct\_sky}: Proportion of sky pixels
\item \texttt{pct\_building}: Proportion of building pixels
\item \texttt{pct\_pavement\_road}: Proportion of pavement/road pixels
\item \texttt{pct\_water}: Proportion of water pixels
\item \texttt{pct\_vehicle\_clutter}: Proportion of vehicle/clutter pixels
\item \texttt{pct\_other}: Proportion of other pixels
\end{itemize}

\begin{figure}[H]
{\centering \pandocbounded{\includegraphics[keepaspectratio]{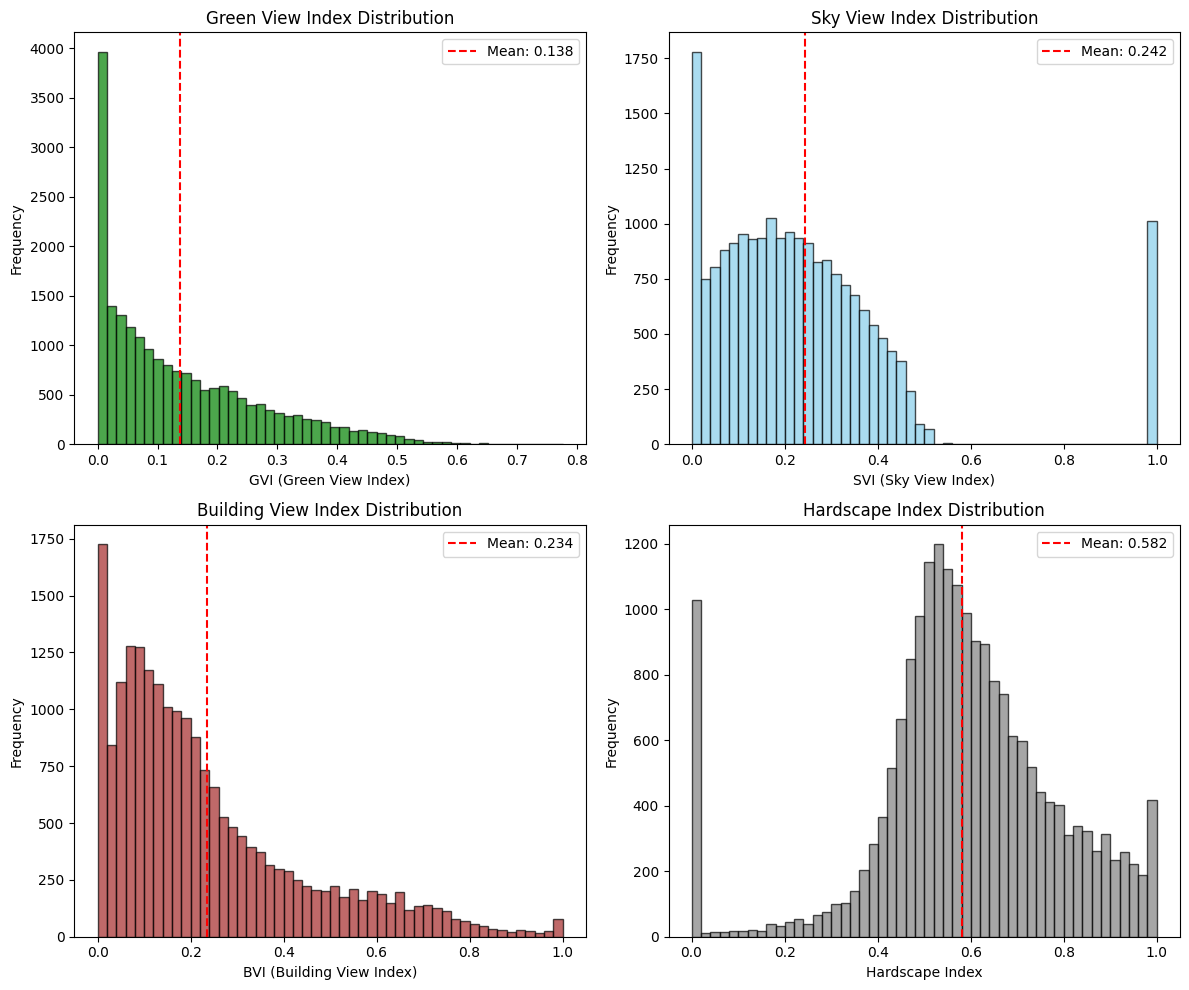}}
}
\caption{GSV-derived streetscape features showing proportions of different urban elements.}
\label{fig:gsv-results}
\end{figure}

\subsection{Raster Extraction}
\label{subsec:extraction}

The collection of Landsat rasters moved through the below workflow in this order:

\begin{enumerate}
\def\labelenumi{\arabic{enumi}.}
\tightlist
\item Create Mosaic Dataset (Data Management)
\item Make Mosaic Layer (Data Management)
\item Cell Statistics (Spatial Analyst)
\item Raster Calculator (Spatial Analyst)
\item Copy Raster (Data Management)
\item Composite Bands (Data Management)
\end{enumerate}

A mosaic dataset was created to combine and stitch together 64 scenes from 2023 thru 2025 during dry season months December thru April.

The different mosaic layers were isolated, with maximum LST and averages for other layers. This was to process individual cell statistics to convert DN raw values and re-scale, and raster calculations to calculate NDVI and alter LST from Kelvin to Celsius. Only the QA\_PIXEL band was acquired through the 5th tool.

Two extraction products were used for the rasters, one for the GSV sample points and second for the entire network nodes. All raster values were extracted for all nodes except for LST, where 0.0037\% of nodes were unaccounted for, decreasing from 28,437 to 28,332 nodes.

\subsection{Model Design and Training}
\label{subsec:model-design}

Two XGBoost models were trained using the \texttt{xgboost} Python library: a full model including raster and GSV features for maximum predictive power, and a deployment model using only raster features for city-wide application \citep{chen2016}.

\begin{Shaded}
\begin{Highlighting}[]
\CommentTok{\# Landsat features: thermal and vegetation indices.}
\NormalTok{LANDSAT\_FEATURES }\OperatorTok{=}\NormalTok{ [}
    \StringTok{"ndvi"}\NormalTok{,}
    \StringTok{"emissivity"}
\NormalTok{]}

\CommentTok{\# PALSAR features: radar backscatter and texture.}
\NormalTok{PALSAR\_FEATURES }\OperatorTok{=}\NormalTok{ [}
    \StringTok{"palsar\_hh\_db"}\NormalTok{,}
    \StringTok{"palsar\_hv\_db"}\NormalTok{,}
    \StringTok{"palsar\_hv\_hh\_ratio"}\NormalTok{,}
    \StringTok{"palsar\_glcm\_contrast"}\NormalTok{,}
    \StringTok{"palsar\_glcm\_homogeneity"}\NormalTok{,}
    \StringTok{"palsar\_glcm\_energy"}
\NormalTok{]}

\CommentTok{\# DSM features: elevation and sky view.}
\NormalTok{DSM\_FEATURES }\OperatorTok{=}\NormalTok{ [}
    \StringTok{"elevation\_m"}\NormalTok{,}
    \StringTok{"sky\_view\_factor"}
\NormalTok{]}

\CommentTok{\# Landcover features.}
\NormalTok{LANDCOVER\_FEATURES }\OperatorTok{=}\NormalTok{ [}
    \StringTok{"landcover\_class"}
\NormalTok{]}

\CommentTok{\# GSV segmentation features: direct superclass percentages.}
\NormalTok{GSV\_SEGMENTATION\_FEATURES }\OperatorTok{=}\NormalTok{ [}
    \StringTok{"pct\_vegetation"}\NormalTok{,}
    \StringTok{"pct\_sky"}\NormalTok{,}
    \StringTok{"pct\_building"}\NormalTok{,}
    \StringTok{"pct\_pavement\_road"}\NormalTok{,}
    \StringTok{"pct\_water"}\NormalTok{,}
    \StringTok{"pct\_vehicle\_clutter"}\NormalTok{,}
    \StringTok{"pct\_other"}
\NormalTok{]}
\end{Highlighting}
\end{Shaded}

\subsection{Feature Selection Rationale}
\label{subsec:features}

Each predictor variable was selected based on its established physical or empirical relationship with land surface temperature:

\textbf{Landsat-Derived Features:}

\begin{itemize}
\item \texttt{ndvi}: The Normalized Difference Vegetation Index quantifies vegetation density. Higher NDVI indicates greater evapotranspiration and shading capacity, which reduces surface temperatures \citep{usgs2024}.
\item \texttt{emissivity}: Surface emissivity determines how efficiently a material radiates absorbed thermal energy. Urban materials (e.g., concrete, asphalt) typically have lower emissivity than vegetated surfaces, affecting the LST retrieval and thermal behavior.
\end{itemize}

\textbf{PALSAR-2 Radar Features:}

\begin{itemize}
\item \texttt{palsar\_hh\_db} and \texttt{palsar\_hv\_db}: SAR backscatter intensity in HH and HV polarizations captures surface roughness and structural characteristics. Built-up areas with vertical structures produce stronger backscatter, serving as proxies for urban density and building mass that store and re-emit heat.
\item \texttt{palsar\_hv\_hh\_ratio}: The cross-polarization ratio distinguishes vegetation (higher HV response due to volume scattering) from built surfaces (dominated by HH) \citep{shimada2014}, providing structural information complementary to optical indices.
\item \texttt{palsar\_glcm\_contrast}, \texttt{palsar\_glcm\_homogeneity}, \texttt{palsar\_glcm\_energy}: Gray-Level Co-occurrence Matrix (GLCM) texture metrics characterize spatial heterogeneity of the urban fabric. High contrast indicates fragmented land cover; homogeneity captures uniformity of surface types---both relate to thermal variability patterns.
\end{itemize}

\textbf{DSM-Derived Features:}

\begin{itemize}
\item \texttt{elevation\_m}: Elevation influences temperature through adiabatic lapse rates and drainage patterns. Lower elevations in HCMC often correspond to denser development and reduced ventilation.
\item \texttt{sky\_view\_factor}: SVF measures the proportion of visible sky hemisphere from a point, approximating urban canyon geometry. Lower SVF indicates taller surrounding structures that trap longwave radiation and reduce nocturnal cooling.
\end{itemize}

\textbf{Land Cover:}

\begin{itemize}
\tightlist
\item \texttt{landcover\_class}: Categorical land use classification directly encodes surface type (water, forest, urban, agriculture), each with distinct thermal properties, albedo, and heat capacity.
\end{itemize}

\textbf{GSV-Derived Streetscape Features:}

\begin{itemize}
\item \texttt{pct\_vegetation}: Street-level vegetation proportion captures micro-scale canopy cover invisible to 30m satellite imagery, directly measuring shade availability at pedestrian height.
\item \texttt{pct\_sky}: Visible sky proportion from street level indicates canyon openness and potential solar exposure---complementing the DSM-derived SVF with human-perspective geometry.
\item \texttt{pct\_building}: Building facade proportion quantifies wall surfaces that absorb and re-radiate heat, contributing to the urban heat island effect at street scale.
\item \texttt{pct\_pavement\_road}: Impervious surface proportion at street level captures heat-absorbing materials with low albedo and no evaporative cooling capacity.
\item \texttt{pct\_water}: Water body visibility indicates proximity to cooling features; water has high heat capacity and provides evaporative cooling.
\item \texttt{pct\_vehicle\_clutter}: Vehicle and street furniture proportion serves as a proxy for traffic density and anthropogenic heat sources.
\item \texttt{pct\_other}: Remaining categories (persons, terrain, miscellaneous objects) provide contextual information about street activity and surface conditions.
\end{itemize}

The XGBoost parameters were carefully selected to balance complexity, learning speed, and regularization to prevent overfitting while capturing complex, non-linear mechanisms that impact LST:

\begin{Shaded}
\begin{Highlighting}[]
\NormalTok{XGB\_PARAMS }\OperatorTok{=}\NormalTok{ \{}
    \StringTok{"n\_estimators"}\NormalTok{: }\DecValTok{500}\NormalTok{,}
    \StringTok{"max\_depth"}\NormalTok{: }\DecValTok{5}\NormalTok{,}
    \StringTok{"learning\_rate"}\NormalTok{: }\FloatTok{0.05}\NormalTok{,}
    \StringTok{"subsample"}\NormalTok{: }\FloatTok{0.8}\NormalTok{,}
    \StringTok{"colsample\_bytree"}\NormalTok{: }\FloatTok{0.8}\NormalTok{,}
    \StringTok{"min\_child\_weight"}\NormalTok{: }\DecValTok{5}\NormalTok{,}
    \StringTok{"reg\_alpha"}\NormalTok{: }\FloatTok{0.5}\NormalTok{,}
    \StringTok{"reg\_lambda"}\NormalTok{: }\FloatTok{2.0}\NormalTok{,}
    \StringTok{"random\_state"}\NormalTok{: RANDOM\_SEED,}
    \StringTok{"n\_jobs"}\NormalTok{: }\OperatorTok{{-}}\DecValTok{1}\NormalTok{,}
    \StringTok{"early\_stopping\_rounds"}\NormalTok{: }\DecValTok{50}
\NormalTok{\}}
\end{Highlighting}
\end{Shaded}

\subsection{Spatial Cross-Validation}
\label{subsec:cv}

To prevent optimistically biased performance estimates from spatial autocorrelation, a leave-one-ward-out spatial cross-validation strategy was implemented. GSV points sampled at 50-meter intervals share the same 30-meter Landsat pixels, so adjacent points in different folds would leak information under standard K-fold CV. By holding out entire wards during each fold, the model must extrapolate to spatially distinct areas, providing realistic generalization estimates.

One ward (An Phú) was reserved as a completely held-out test set, never seen during any stage of training or hyperparameter tuning. This provides an unbiased estimate of true generalization performance.

\subsection{Node Prediction and Routing}
\label{subsec:routing-implementation}

A patchwork approach was pursued, using the full model (with GSV features) to predict within wards of interest and the deployment model (raster-only) to predict outside of those wards where GSV imagery is unavailable.

After generating predictions for network nodes, a prediction raster was derived from them and interpolated to create a hybrid cost surface. The raster resolution was 0.0001 degrees and 2,266 × 1,409 pixel resolution clipped to the area of interest encompassing the districts, and a Gaussian blur (sigma = 4) was applied to smooth out interpolation artifacts.

Dijkstra's algorithm was implemented \citep{dijkstra1959}, assigning heat edge costs by combining normalized length and temperature to support three route types with tunable heat penalty and reward parameters: shortest, coolest with a temperature penalty, and hottest with an inverted temperature penalty to reward.

\begin{Shaded}
\begin{Highlighting}[]
\CommentTok{\# Heat weights.}
\ControlFlowTok{for}\NormalTok{ u, v, data }\KeywordTok{in}\NormalTok{ G\_undirected.edges(data }\OperatorTok{=} \VariableTok{True}\NormalTok{):}
\NormalTok{    avg\_lst }\OperatorTok{=}\NormalTok{ data[}\StringTok{"avg\_lst"}\NormalTok{]}
\NormalTok{    length  }\OperatorTok{=}\NormalTok{ data[}\StringTok{"length"}\NormalTok{]}

    \CommentTok{\# Normalize temperature to [0, 1].}
\NormalTok{    temp\_norm }\OperatorTok{=}\NormalTok{ (avg\_lst }\OperatorTok{{-}}\NormalTok{ lst\_min) }\OperatorTok{/}\NormalTok{ lst\_range}

    \CommentTok{\# Normalize length relative to mean edge length.}
\NormalTok{    length\_norm }\OperatorTok{=}\NormalTok{ length }\OperatorTok{/}\NormalTok{ len\_mean}

    \CommentTok{\# Tune lambda.}
\NormalTok{    lambda\_cool }\OperatorTok{=} \FloatTok{10.0} \CommentTok{\# How much to penalize heat.}
\NormalTok{    lambda\_hot  }\OperatorTok{=} \FloatTok{10.0} \CommentTok{\# How much to reward heat.}

    \CommentTok{\# Cool Cost: shorter + cooler (high temperature = high penalty).}
\NormalTok{    data[}\StringTok{"cool\_cost"}\NormalTok{] }\OperatorTok{=}\NormalTok{ length\_norm }\OperatorTok{+}\NormalTok{ lambda\_cool }\OperatorTok{*}\NormalTok{ temp\_norm}

    \CommentTok{\# Hot Cost: shorter + hotter (invert temp\_norm).}
\NormalTok{    data[}\StringTok{"hot\_cost"}\NormalTok{] }\OperatorTok{=}\NormalTok{ length\_norm }\OperatorTok{+}\NormalTok{ lambda\_hot }\OperatorTok{*}\NormalTok{ (}\FloatTok{1.0} \OperatorTok{{-}}\NormalTok{ temp\_norm)}
\end{Highlighting}
\end{Shaded}

\begin{Shaded}
\begin{Highlighting}[]
\CommentTok{\# Routing helpers.}
\KeywordTok{def}\NormalTok{ get\_hottest\_route(G, start, end):}
    \CommentTok{"""Shortest{-}ish but biased toward hottest edges."""}
    \ControlFlowTok{try}\NormalTok{:}
\NormalTok{        path }\OperatorTok{=}\NormalTok{ nx.dijkstra\_path(G, start, end, weight }\OperatorTok{=} \StringTok{"hot\_cost"}\NormalTok{)}
\NormalTok{        cost }\OperatorTok{=}\NormalTok{ nx.dijkstra\_path\_length(G, start, end, weight }\OperatorTok{=} \StringTok{"hot\_cost"}\NormalTok{)}
\NormalTok{        dist }\OperatorTok{=}\NormalTok{ nx.path\_weight(G, path, weight }\OperatorTok{=} \StringTok{"length"}\NormalTok{)}
        \ControlFlowTok{return}\NormalTok{ path, cost, dist}
    \ControlFlowTok{except}\NormalTok{ nx.NetworkXNoPath:}
        \ControlFlowTok{return} \VariableTok{None}\NormalTok{, }\VariableTok{None}\NormalTok{, }\VariableTok{None}

\KeywordTok{def}\NormalTok{ get\_coolest\_route(G, start, end):}
    \CommentTok{"""Shortest{-}ish but biased toward coolest edges."""}
    \ControlFlowTok{try}\NormalTok{:}
\NormalTok{        path }\OperatorTok{=}\NormalTok{ nx.dijkstra\_path(G, start, end, weight }\OperatorTok{=} \StringTok{"cool\_cost"}\NormalTok{)}
\NormalTok{        cost }\OperatorTok{=}\NormalTok{ nx.dijkstra\_path\_length(G, start, end, weight }\OperatorTok{=} \StringTok{"cool\_cost"}\NormalTok{)}
\NormalTok{        dist }\OperatorTok{=}\NormalTok{ nx.path\_weight(G, path, weight }\OperatorTok{=} \StringTok{"length"}\NormalTok{)}
        \ControlFlowTok{return}\NormalTok{ path, cost, dist}
    \ControlFlowTok{except}\NormalTok{ nx.NetworkXNoPath:}
        \ControlFlowTok{return} \VariableTok{None}\NormalTok{, }\VariableTok{None}\NormalTok{, }\VariableTok{None}

\KeywordTok{def}\NormalTok{ get\_shortest\_route(G, start, end):}
    \CommentTok{"""}
\CommentTok{    Returns shortest walking path.}
\CommentTok{    """}
    \ControlFlowTok{try}\NormalTok{:}
\NormalTok{        path }\OperatorTok{=}\NormalTok{ nx.shortest\_path(G, start, end, weight }\OperatorTok{=} \StringTok{"length"}\NormalTok{)}
\NormalTok{        dist }\OperatorTok{=}\NormalTok{ nx.path\_weight(G, path, weight }\OperatorTok{=} \StringTok{"length"}\NormalTok{)}
        \ControlFlowTok{return}\NormalTok{ path, dist}
    \ControlFlowTok{except}\NormalTok{ nx.NetworkXNoPath:}
        \ControlFlowTok{return} \VariableTok{None}\NormalTok{, }\VariableTok{None}
\end{Highlighting}
\end{Shaded}

\section{Results}
\label{sec:results}

\subsection{Model Performance}
\label{subsec:performance}

\begin{figure}[H]
{\centering \pandocbounded{\includegraphics[keepaspectratio]{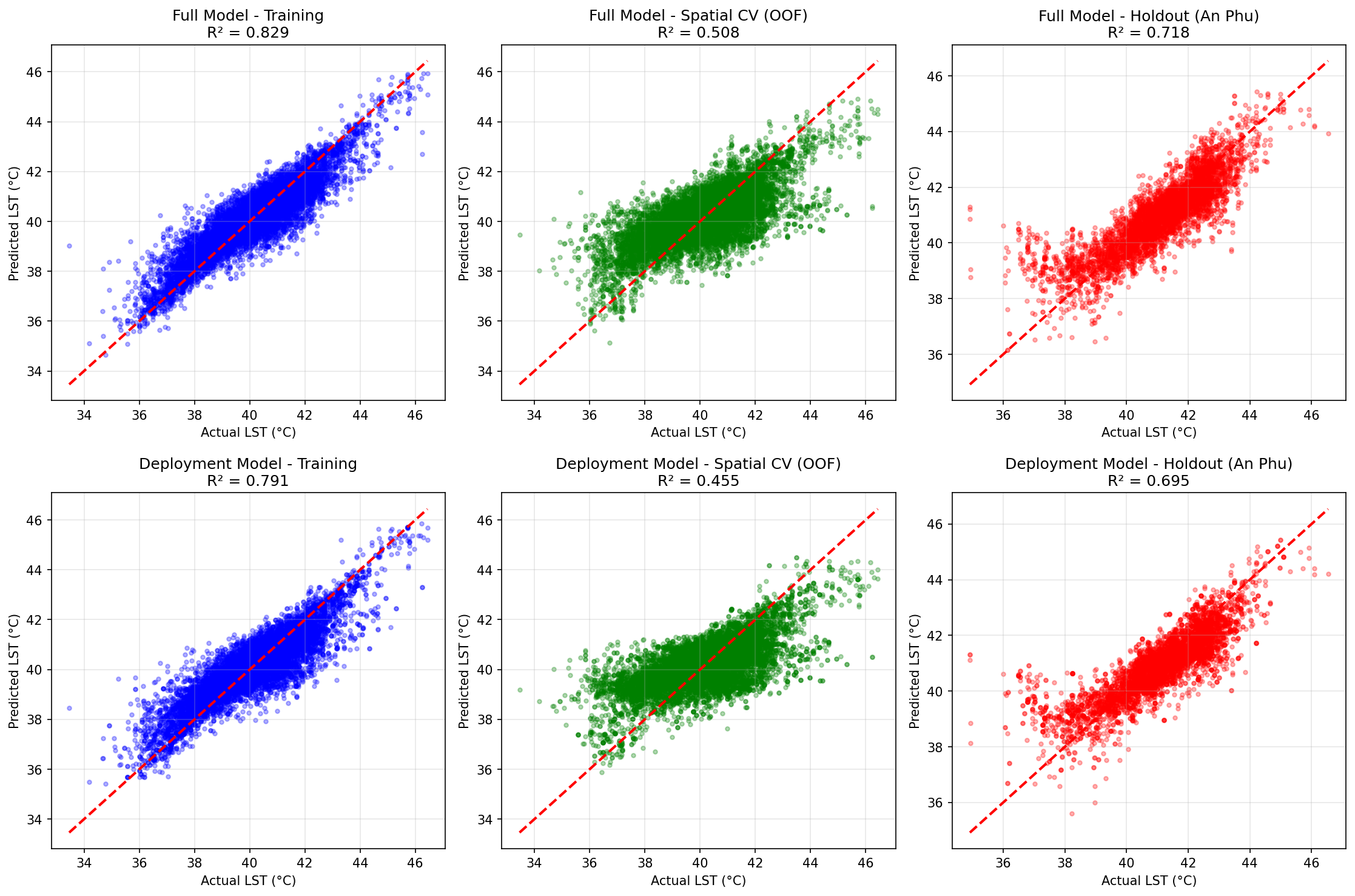}}
}
\caption{Scatter plot showing predicted vs. actual LST values for the holdout ward.}
\label{fig:actual-predicted}
\end{figure}

\begin{figure}[H]
{\centering \pandocbounded{\includegraphics[keepaspectratio]{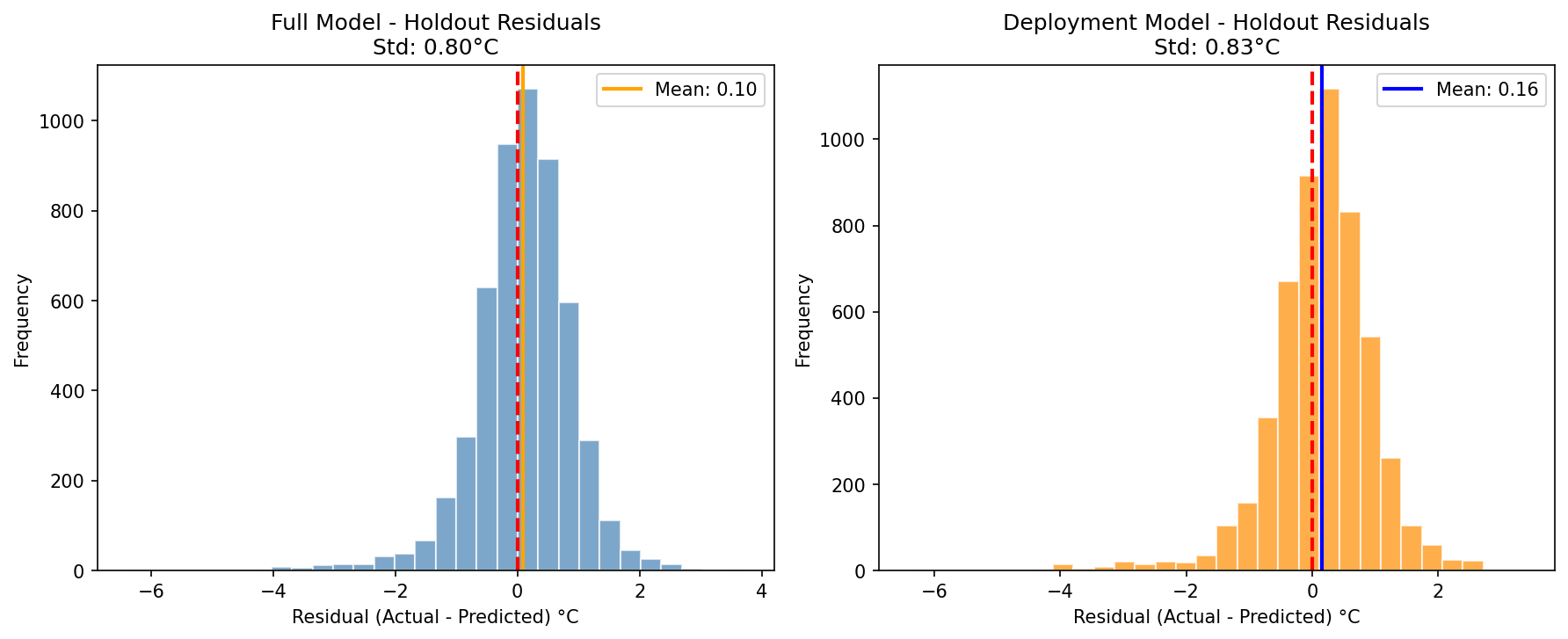}}
}
\caption{Distribution of prediction residuals showing model bias characteristics.}
\label{fig:residuals}
\end{figure}

Model performance was evaluated using three complementary metrics: training set performance, spatial cross-validation (leave-one-ward-out), and holdout ward (An Phú) performance.

\begin{longtable}[]{@{}lll@{}}
\caption{Model performance comparison between full model (with GSV features) and deployment model (raster-only).}\label{tab:performance}\\
\toprule\noalign{}
Metric & Full Model & Deployment Model \\
\midrule\noalign{}
\endfirsthead
\toprule\noalign{}
Metric & Full Model & Deployment Model \\
\midrule\noalign{}
\endhead
\bottomrule\noalign{}
\endlastfoot
Features & 18 & 11 \\
Training RMSE & 0.6203 & 0.6844 \\
Training R² & 0.8285 & 0.7913 \\
Spatial CV RMSE & 1.0510 ± 0.2898 & 1.1061 ± 0.2755 \\
Spatial CV R² & 0.5079 ± 0.2757 & 0.4549 ± 0.2688 \\
Holdout RMSE & 0.8093 & 0.8422 \\
Holdout R² & 0.7180 & 0.6946 \\
Holdout MAE & 0.5909 & 0.6134 \\
\end{longtable}

\begin{figure}[H]
{\centering \pandocbounded{\includegraphics[keepaspectratio]{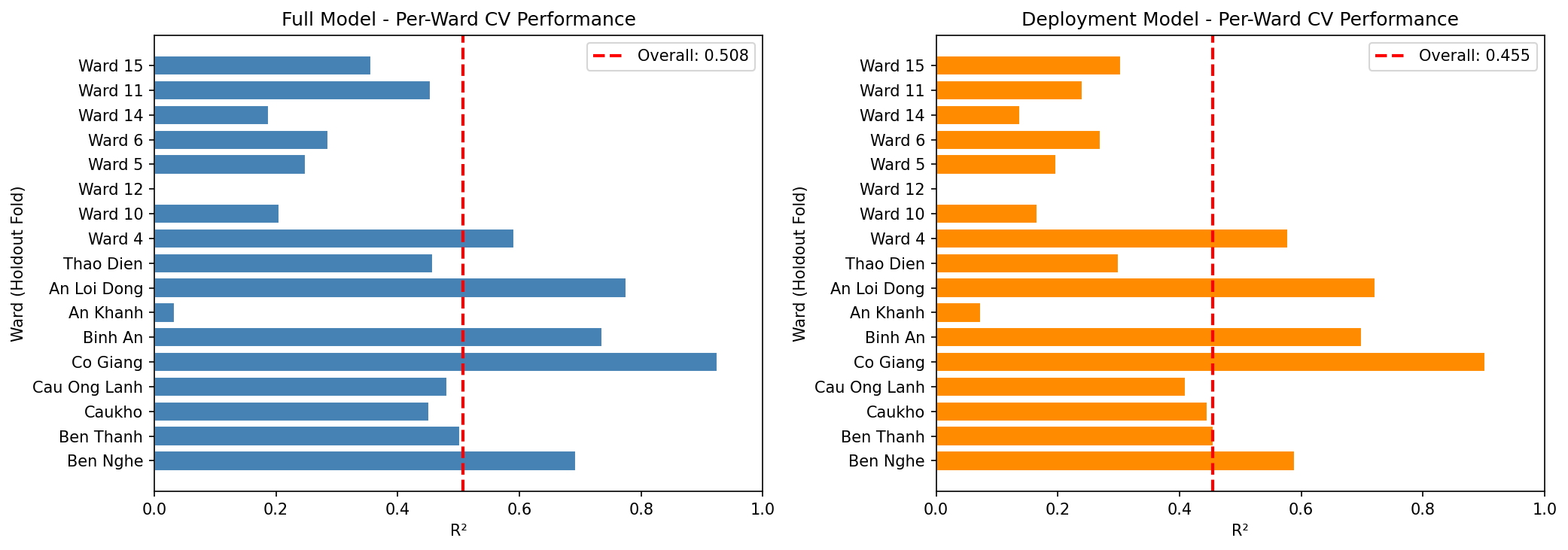}}
}
\caption{Cross-validation performance varies by ward, reflecting differences in urban morphology.}
\label{fig:ward-cv}
\end{figure}

The spatial cross-validation reveals substantial inter-ward variability (± 0.27 R²), indicating that some wards are considerably harder to predict than others. The holdout ward (An Phú) achieved higher R² than the spatial CV average, suggesting it shares similar characteristics with the training wards.

\subsection{Overfitting Assessment}
\label{subsec:overfitting}

\begin{longtable}[]{@{}lll@{}}
\caption{Comparison of training, cross-validation, and holdout performance to assess model overfitting.}\label{tab:overfitting}\\
\toprule\noalign{}
Metric & Full Model & Deployment Model \\
\midrule\noalign{}
\endfirsthead
\toprule\noalign{}
Metric & Full Model & Deployment Model \\
\midrule\noalign{}
\endhead
\bottomrule\noalign{}
\endlastfoot
Training R² & 0.8285 & 0.7913 \\
Spatial CV R² & 0.5079 & 0.4549 \\
Holdout R² & 0.7180 & 0.6946 \\
Train-Holdout Gap & 0.1105 & 0.0967 \\
\end{longtable}

The train-holdout gap of approximately 0.10 indicates moderate but acceptable overfitting for a spatial prediction task. The larger train-CV gap ($\sim$0.32) reflects inter-ward heterogeneity rather than classical overfitting---different wards have different predictability based on their urban morphology.

\subsection{GSV Feature Contribution}
\label{subsec:gsv-contribution}

The contribution of GSV-derived streetscape features was quantified by comparing the full model (with GSV) to the deployment model (raster-only):

\begin{longtable}[]{@{}ll@{}}
\caption{Improvement in model performance attributable to GSV-derived features.}\label{tab:gsv-improvement}\\
\toprule\noalign{}
Metric & Improvement \\
\midrule\noalign{}
\endfirsthead
\toprule\noalign{}
Metric & Improvement \\
\midrule\noalign{}
\endhead
\bottomrule\noalign{}
\endlastfoot
Spatial CV R² Improvement & +0.0530 \\
Holdout R² Improvement & +0.0234 \\
\end{longtable}

GSV features provide modest but meaningful improvement, confirming that street-level imagery adds predictive value beyond satellite data alone.

\subsection{Feature Importance}
\label{subsec:feature-importance}

\begin{figure}[H]
{\centering \pandocbounded{\includegraphics[keepaspectratio]{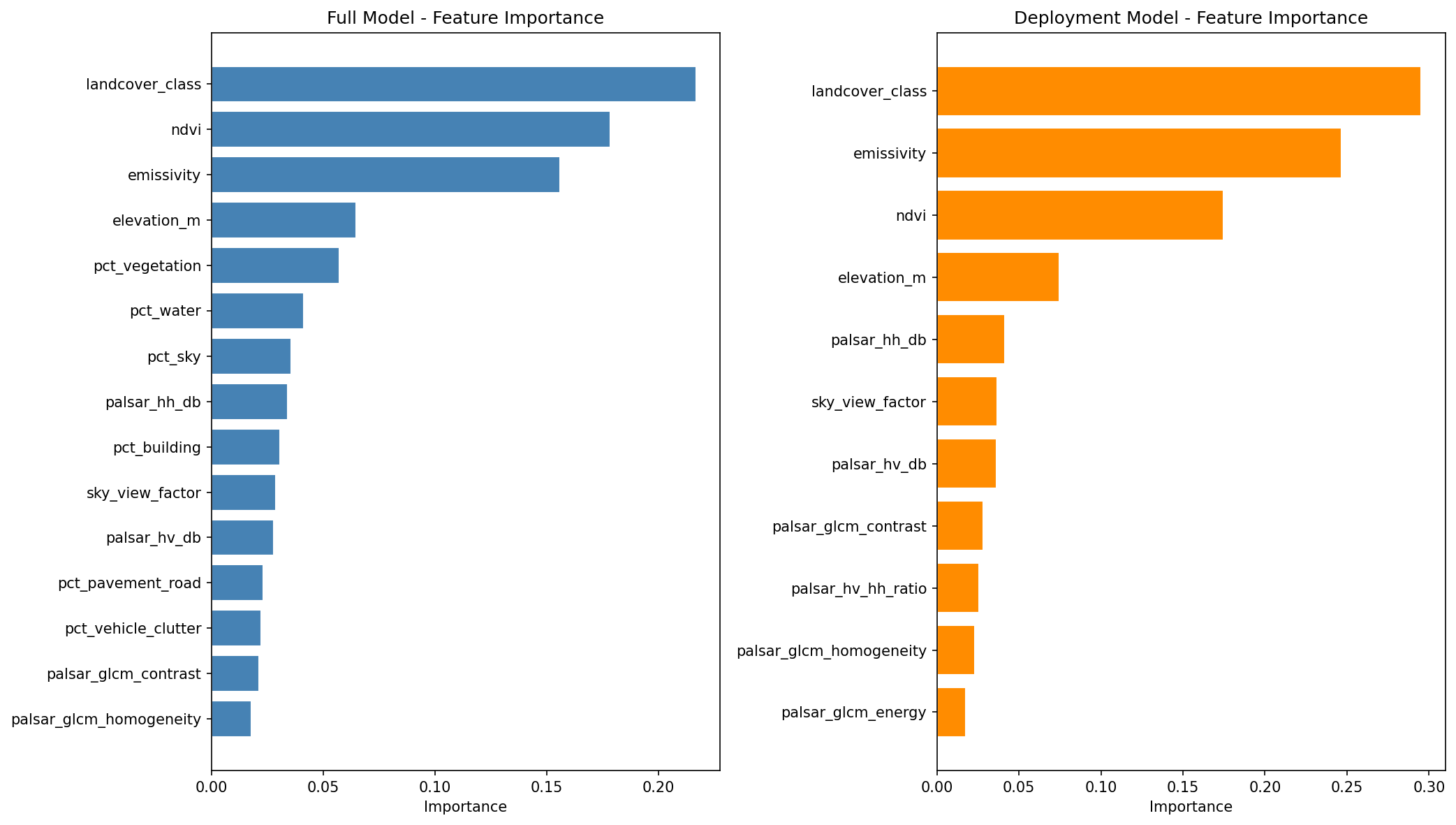}}
}
\caption{Feature importance rankings from XGBoost models showing dominant predictors of LST.}
\label{fig:importance}
\end{figure}

\begin{longtable}[]{@{}lll@{}}
\caption{Top 10 features by importance score in both full and deployment models.}\label{tab:top-features}\\
\toprule\noalign{}
\textbf{Feature} & \textbf{Full Model} & \textbf{Deployment Model} \\
\midrule\noalign{}
\endfirsthead
\toprule\noalign{}
\textbf{Feature} & \textbf{Full Model} & \textbf{Deployment Model} \\
\midrule\noalign{}
\endhead
\bottomrule\noalign{}
\endlastfoot
landcover\_class & 0.2169 & 0.2953 \\
ndvi & 0.1787 & 0.1746 \\
emissivity & 0.1560 & 0.2467 \\
elevation\_m & 0.0648 & 0.0746 \\
pct\_vegetation & 0.0572 & --- \\
pct\_water & 0.0413 & --- \\
pct\_sky & 0.0355 & --- \\
palsar\_hh\_db & 0.0342 & 0.0416 \\
pct\_building & 0.0305 & --- \\
sky\_view\_factor & 0.0289 & 0.0366 \\
\end{longtable}

The dominant predictors were consistent across both models. Landcover class, NDVI, and emissivity ranked highest, together accounting for over 55\% of predictive power. In the full model, GSV-derived streetscape variables (vegetation, water, sky percentages) contributed additional explanatory signal, ranking within the top 10 features.

\subsection{Node Prediction Coverage}
\label{subsec:coverage}

\begin{longtable}[]{@{}llll@{}}
\caption{Summary statistics for predicted LST across all network nodes.}\label{tab:lst-stats}\\
\toprule\noalign{}
Minimum & Mean & Maximum & Standard Deviation \\
\midrule\noalign{}
\endfirsthead
\toprule\noalign{}
Minimum & Mean & Maximum & Standard Deviation \\
\midrule\noalign{}
\endhead
\bottomrule\noalign{}
\endlastfoot
35.44°C & 40.84°C & 46.28°C & 1.76°C \\
\end{longtable}

\begin{figure}[H]
{\centering \pandocbounded{\includegraphics[keepaspectratio]{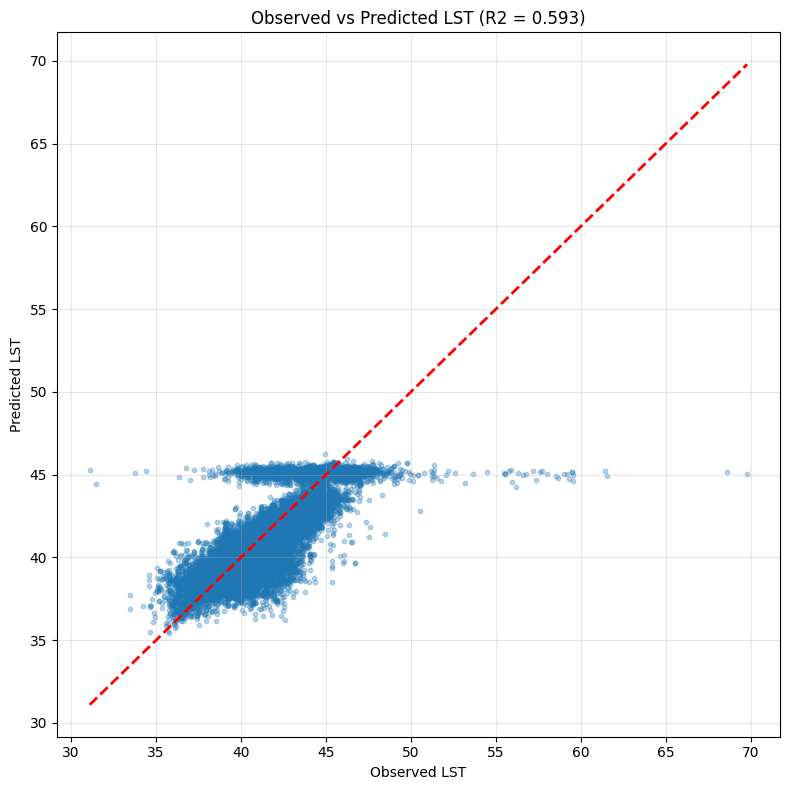}}
}
\caption{Scatter plot comparing observed Landsat LST with model predictions at network nodes.}
\label{fig:lst-scatter}
\end{figure}

\begin{figure}[H]
{\centering \pandocbounded{\includegraphics[keepaspectratio]{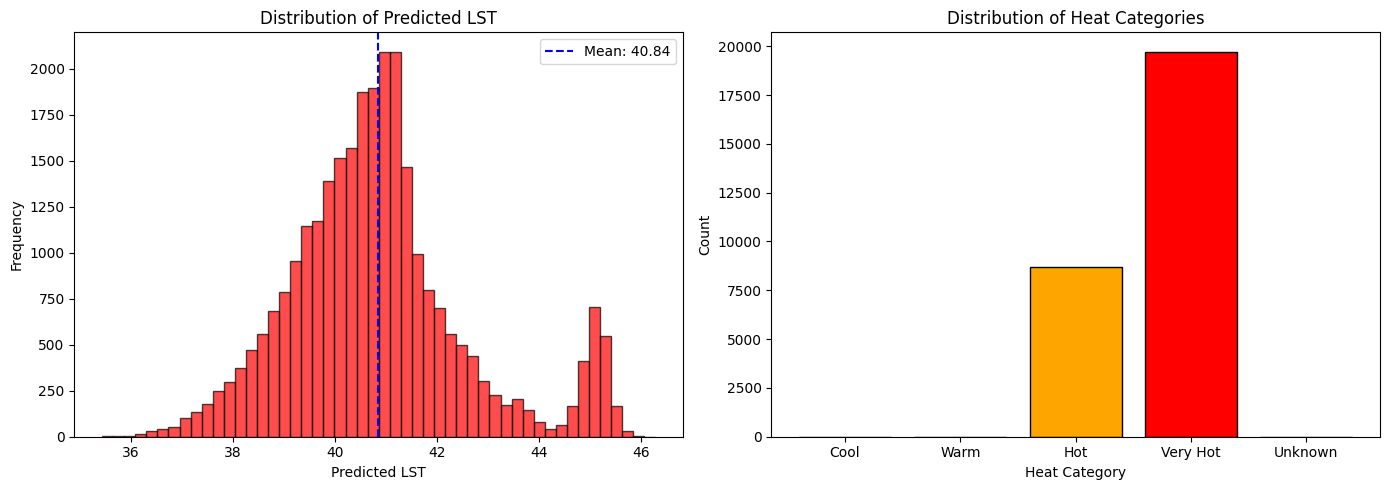}}
}
\caption{Distribution of predicted LST values across the pedestrian network.}
\label{fig:lst-dist}
\end{figure}

\begin{figure}[H]
{\centering \pandocbounded{\includegraphics[keepaspectratio]{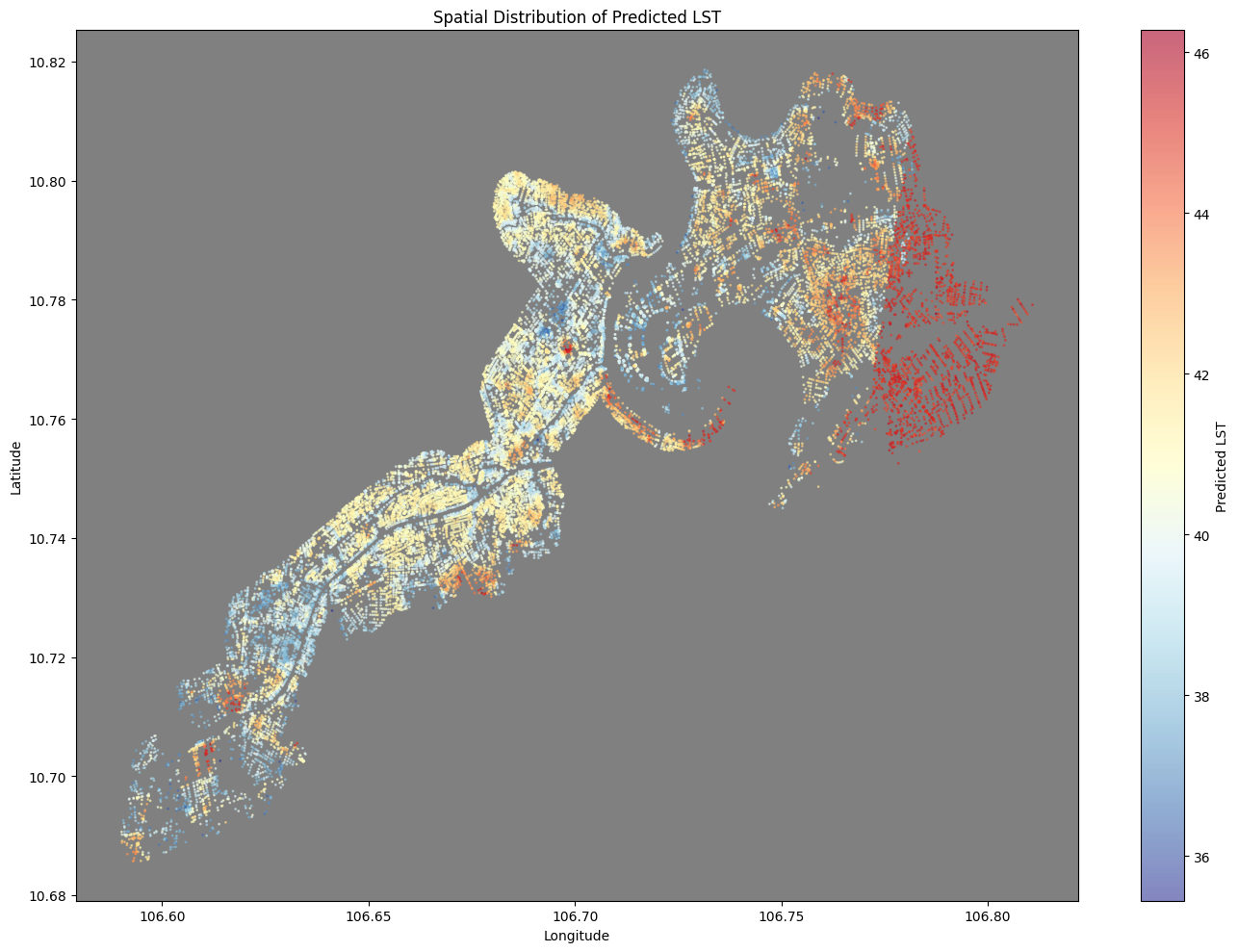}}
}
\caption{Spatial map of predicted LST showing thermal heterogeneity across the study area.}
\label{fig:lst-map}
\end{figure}

This mean-reverting behavior has critical implications for applying the model to environments outside the training distribution, particularly areas with significant open space, non-tree greenery like grass fields, or non-clustered buildings. The model was trained primarily on the complex street canyon geometry of HCMC's central districts. It may struggle to fully capture the cooling effect of large, continuous patches of greenery and open spaces because these structures were not the dominant features in the training data's morphology. Consequently, it is likely to over-predict the LST in these large, cool areas, pushing the prediction closer to the mean and failing to accurately capture the full range of low temperatures achievable in those settings.

\subsection{Routing Outcomes}
\label{subsec:routing-results}

The coolest route increases travel distance but reduces both mean and peak exposure, demonstrating tangible potential for heat-resilient mobility guidance. The hottest route identifies corridors of maximum heat exposure---priority candidates for infrastructure intervention.

\begin{figure}[H]
{\centering \pandocbounded{\includegraphics[keepaspectratio]{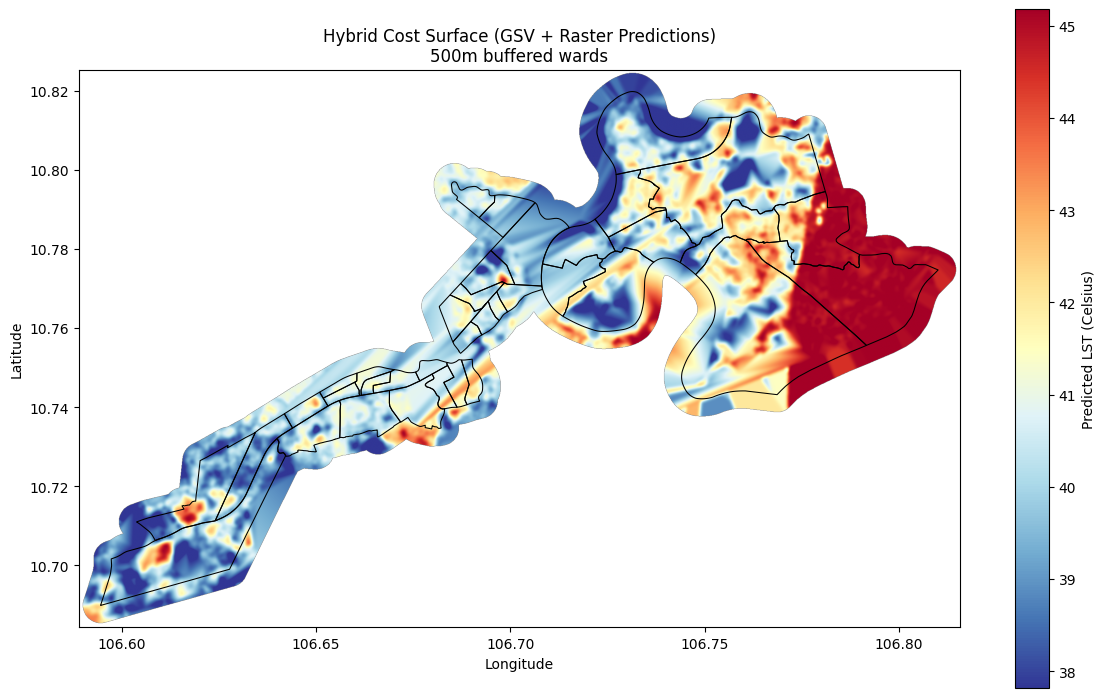}}
}
\caption{Hybrid cost surface raster combining predicted LST with distance for routing optimization.}
\label{fig:hybrid-raster}
\end{figure}

\begin{figure}[H]
{\centering \pandocbounded{\includegraphics[keepaspectratio]{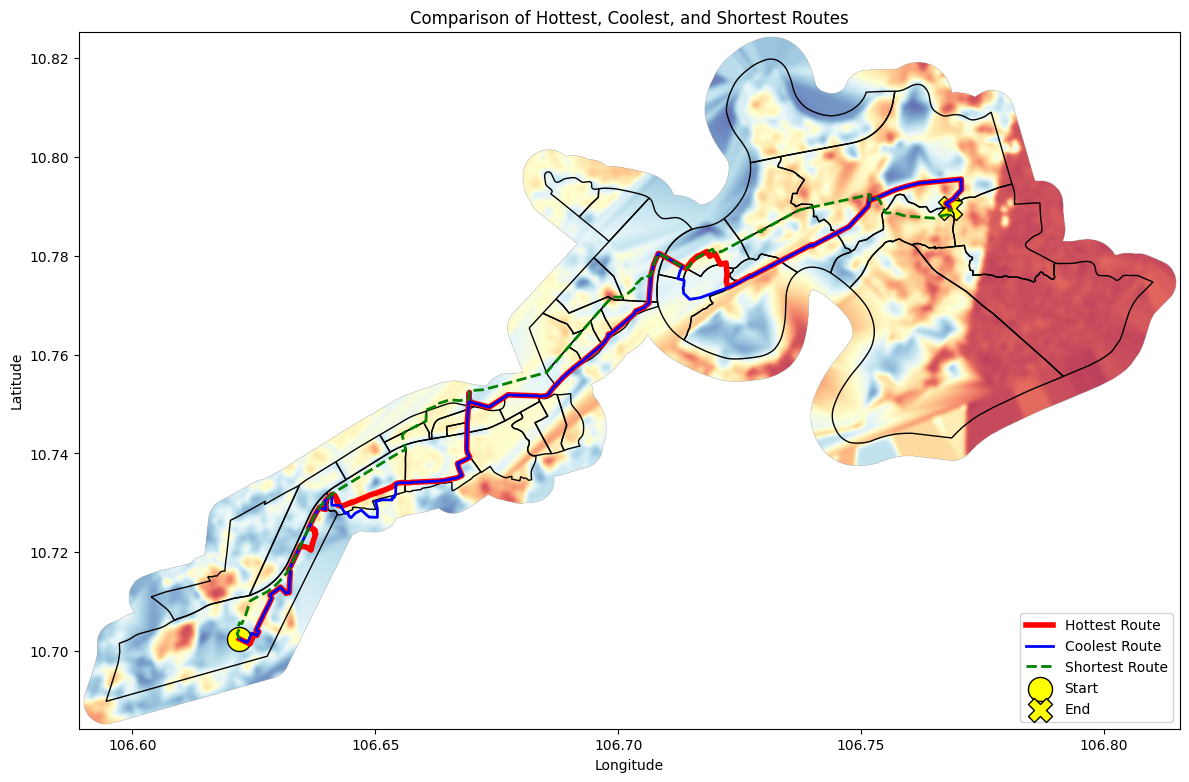}}
}
\caption{Comparison of three routing strategies: shortest path, coolest route, and hottest route.}
\label{fig:routes}
\end{figure}

\begin{figure}[H]
{\centering \pandocbounded{\includegraphics[keepaspectratio]{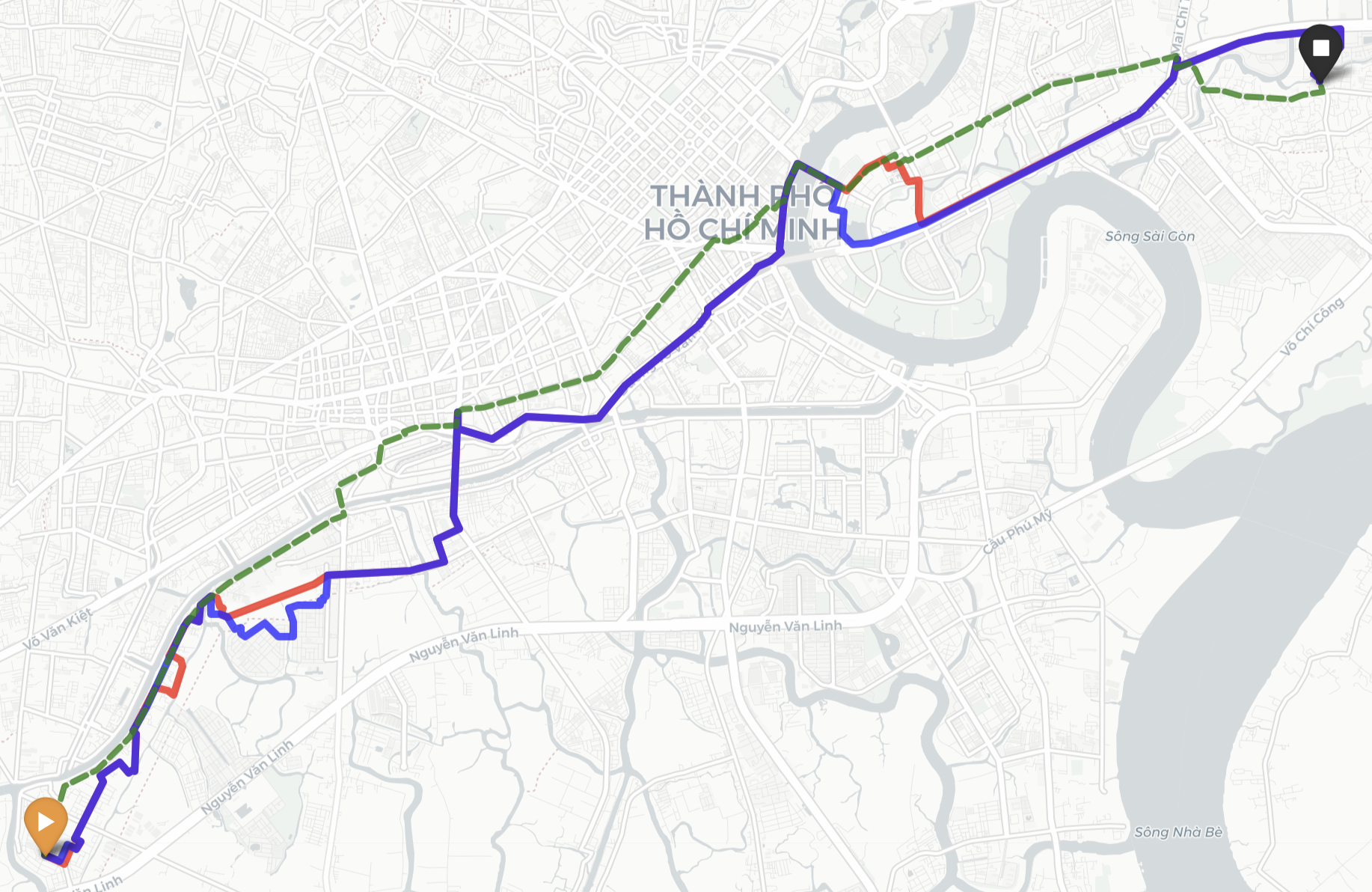}}
}
\caption{Cartographic visualization of heat-aware routing results in urban context.}
\label{fig:carto}
\end{figure}

\begin{longtable}[]{@{}llll@{}}
\caption{Comparison of routing outcomes showing distance and thermal exposure trade-offs.}\label{tab:routing}\\
\toprule\noalign{}
Route & Distance (km) & Average LST & Maximum LST \\
\midrule\noalign{}
\endfirsthead
\toprule\noalign{}
Route & Distance (km) & Average LST & Maximum LST \\
\midrule\noalign{}
\endhead
\bottomrule\noalign{}
\endlastfoot
Shortest & 22.00 & 40.63°C & 45.41°C \\
Coolest & 27.83 & 40.20°C & 43.95°C \\
Hottest & 27.38 & 40.40°C & 43.95°C \\
\end{longtable}

The coolest route adds 5.83 km (26.5\% distance penalty) to reduce average temperature by 0.43°C. This trade-off highlights why infrastructure investment matters more than individual route choice---residents should not bear a 26\% distance penalty to marginally reduce heat exposure.

\section{Discussion and Limitations}
\label{sec:discussion}

Hot Hẻm demonstrates a scalable approach for integrating human-scale streetscape morphology with city-scale remote sensing to operationalize pedestrian heat risk. The deployment model achieves R² of 0.69 on the held-out An Phú ward, with predictions typically within 0.61°C (MAE) of observed LST, demonstrating robust generalization to unseen areas. However, there are significant limitations for the future:

\textbf{Spatial Transferability:} Performance varies significantly by ward (CV std = ±0.27 R²). Some areas with unique urban morphology are harder to predict, and the model may underperform in wards that differ substantially from the training distribution \citep{middel2019}.

\textbf{Temporal Mismatch:} GSV images were captured at various times over several years, while Landsat composites represent dry-season 2023--2025 maximum temperatures. Street conditions (e.g., tree canopy, construction) may have changed between GSV capture and satellite observation.

\textbf{Feature Redundancy:} The originally computed GVI, SVI, and BVI indices were identical to their corresponding superclass percentages (pct\_vegetation, pct\_sky, pct\_building) and were removed from the final model to avoid redundancy.

\textbf{Sky View Factor Limitations:} The sky view factor derived from the 30m terrain DSM captures topographic effects but does not fully represent urban canyon geometry at the street level.

\textbf{Heat Category Calibration:} The current heat-category thresholds skew heavily toward "Hot / Very Hot," suggesting that categorical calibration (e.g., quantile-based or health-relevant thresholds) should be refined prior to policy-facing deployment.

\section{Conclusion}
\label{sec:conclusion}

This project delivers a reproducible, multi-scale GeoAI pipeline for heat-weighted pedestrian routing in Ho Chi Minh City. By combining GSV-derived segmentation indices with Landsat thermal variables, JAXA SAR structure, and DSM terrain context, the framework achieves strong predictive accuracy (holdout R² = 0.70, MAE = 0.6°C) and enables practical routing alternatives that identify heat exposure corridors.

The key insight is methodological: rather than helping individuals escape heat, the hottest route optimization identifies where pedestrians suffer most, providing municipalities with actionable data for infrastructure intervention. The 26\% distance penalty imposed by the coolest route demonstrates that heat avoidance should not be framed as individual responsibility---it is a systemic infrastructure challenge requiring public investment. It should be noted that GSV imagery contains copyright restrictions forbidding their implementations in building applications, so granular streetscape imagery would need to be manually obtained or downloaded from open-source material.

Future extensions should include multi-ward holdout testing, threshold calibration using health-relevant cutoffs, multi-season or diurnal modeling, weather data, uncertainty-aware routing, and Meta’s tree canopy height data \citep{tolan2023}, and Global Building Atlas’ 3D dataset \citep{zhu2025} to further strengthen real-world applicability.

\section*{Acknowledgements}

Completed as a requirement in MUSA 6950-001: AI for Urban Sustainability, taught by Dr. Xiaojiang Li at the University of Pennsylvania Stuart Weitzman School of Design's Master of Urban Spatial Analytics program.

\section*{Code Availability}

GitHub Repository: \url{https://github.com/tess-vu/hot-hem}

% Bibliography using BibTeX.
\bibliography{references}

\end{document}